\documentclass[pdflatex,sn-mathphys-num]{sn-jnl}

\usepackage{graphicx}%
\usepackage{multirow}%
\usepackage{amsmath,amssymb,amsfonts}%
\usepackage{amsthm}%
\usepackage{mathrsfs}%
\usepackage[title]{appendix}%
\usepackage{xcolor}%
\usepackage{textcomp}%
\usepackage{manyfoot}%
\usepackage{booktabs}%
\usepackage{algorithm}%
\usepackage{algorithmicx}%
\usepackage{algpseudocode}%
\usepackage{listings}%

\theoremstyle{thmstyleone}%
\theoremstyle{thmstyletwo}%

\theoremstyle{thmstylethree}%

\begin{document}

\title{Ethical considerations of use of hold-out sets in clinical prediction model management}

\author[1]{\fnm{Louis} \sur{Chislett}}\email{louis.chislett@ed.ac.uk}

\author[2,3]{\fnm{Louis JM} \sur{Aslett}}\email{louis.aslett@durham.ac.uk}

\author[3]{\fnm{Alisha R} \sur{Davies}}\email{alisha.davies@wales.nhs.uk}

\author[1,3]{\fnm{Catalina A} \sur{Vallejos}}\email{catalina.vallejos@ed.ac.uk}

\author*[2]{\fnm{James} \sur{Liley}}\email{james.liley@durham.ac.uk}

\affil*[1]{\orgdiv{MRC Human Genetics Unit}, \orgname{University of Edinburgh}, \orgaddress{\street{Crewe Rd S}, \city{Edinburgh}, \postcode{EH4 2XU}, \country{United Kingdom}}}

\affil[2]{\orgdiv{Department of Mathematical Sciences}, \orgname{Durham University}, \orgaddress{\street{Stockton Rd}, \city{Durham}, \postcode{DH1 3LE}, \country{United Kingdom}}}

\affil[3]{\orgdiv{The Alan Turing Institute}, \orgaddress{\street{96 Euston Rd}, \city{London}, \postcode{NW1 2DB}, \country{United Kingdom}}}

\abstract{Clinical prediction models are statistical or machine learning models used to quantify the risk of a certain health outcome using patient data. These can then inform potential interventions on patients, causing an effect called performative prediction: predictions inform interventions which influence the outcome they were trying to predict, leading to a potential underestimation of risk in some patients if a model is updated on this data. One suggested resolution to this is the use of hold-out sets, in which a set of patients do not receive model derived risk scores, such that a model can be safely retrained. We present an overview of clinical and research ethics regarding potential implementation of hold-out sets for clinical prediction models in health settings. We focus on the ethical principles of beneficence, non-maleficence, autonomy and justice. We also discuss informed consent, clinical equipoise, and truth-telling. We present illustrative cases of potential hold-out set implementations and discuss statistical issues arising from different hold-out set sampling methods. We also discuss differences between hold-out sets and randomised control trials, in terms of ethics and statistical issues. Finally, we give practical recommendations for researchers interested in the use hold-out sets for clinical prediction models. }

\keywords{Performative Prediction, Hold-out sets, Clinical Prediction Models}

\maketitle

\section{Introduction}\label{introduction}

One typical goal of machine learning is to predict an outcome $Y$ given a set of covariates (features) $X$\cite{Hastie2001TheLearning}. Here we focus on binary outcomes, where $Y$ denotes whether an event of interest occurs ($Y=1$) or not ($Y=0$). Risk scores are often used to estimate the probability of observing the event $Y = 1$ given the values of $X$. In healthcare applications, this is commonly referred to as a clinical prediction model (CPM), where $X$ captures the patient-level information and $Y$ would represent a health outcome being present or occurring in the future \cite{Topol2019High-performanceIntelligence}. It is worth noting that models which aim to predict a probability of future outcome are referred to as ``prognostic rules'', while those that aim to predict the probability of an intervention being successful are called ``prescriptive rules''\cite{Cowley2019MethodologicalLiterature}. This paper primarily deals with prognostic rules. Thus the output of a CPM in our case is a risk score, that is $Pr(Y=1 | X)$. For example, the EuroSCORE II CPM predicts whether a patient will die prior to hospital discharge after cardiac surgery using the patient's age, gender and existing medical conditions as input covariates\cite{Nashef2012EuroscoreII}.

A fundamental issue in the development and application of CPMs is their accuracy and how this evolves over time. If the distribution of $(X,Y)$ changes (often called ``drift''), then the risk scores may become biased, leading to less accurate predictions\cite{Zliobaite2010LearningOverview}. For example, one study showed that drift affecting model accuracy occurred in several machine learning methods when predicting 30-day mortality after hospital admission \cite{Davis2020DetectionUpdating}. 

Reasons for drift can be diverse and can act simultaneously\cite{Finlayson2021TheIntelligence}. These can include changes in the distribution of $Y$ (e.g.~increase prevalence of a disease), 
changes in relationships between $X$ and $Y$ (e.g.~a risk factor becoming less predictive), changes in the distribution of $X$ (e.g.~an aging population, if age is a predictor), as well as ``performative prediction'' effects\cite{Perdomo2020PerformativePrediction}\cite{Toll2008ValidationReview}. Performative prediction is a potentially understudied concept in which a risk score influences the distribution of $(X,Y)$ through interventions made based on its predictions. When used in this way, CPMs inform what are referred to as Clinical Decision Support Systems \cite{Sutton2020AnSuccess}. In a health context, interventions made based on risk scores distributed to patients or doctors may induce some performative effect: individuals with high risk scores may be prioritised for intervention which, if successful, may reduce their risk - thus changing the relationship between $X$ and $Y$. For example with QRISK3, a model which predicts 10-year risk of cardiovascular disease, a high risk score can inform an intervention such as a prescription of statins for the patient \cite{Hippisley-Cox2017DevelopmentStudy}. If the model were to be retrained using post-intervention data as input, similar patients would receive risk scores which are underestimations of their true risk\cite{Liley2021ModelBias}. Performative effects can be amplified by having a more effective model or intervention, resulting in models becoming ``victims of their own success''\cite{Lenert2019PrognosticUnless...}.

There is no consensus within the field on how to deal with a performative effect in a prediction-intervention system. The natural response to drift is to re-train (update) the model using more recent data. However, as illustrated above, this is not optimal in the presence of performative effects. One method which has been proposed is to ``hold-out'' a set of individuals who do not receive risk score guided interventions, on which a model can be retrained\cite{Liley2021ModelBias}. Hold-out sets are similar to control arms in clinical trials, but the goal is to mitigate the effects of performative prediction when updating a CPM rather than to measure the effect of the CPM itself (such as whether a measured decrease in patients with $Y=1$ occurs). While hold-out sets may be trivial to implement in other contexts where the prediction does not affect a person's well-being (e.g.~a social media platform which uses a model to predict whether someone will click an advertisement), this is not the case with CPMs. If hold-out sets are to be a solution to the performative prediction problem in CPMs, there are a number of ethical, practical and statistical issues which need to be considered. We consider these issues in the context of different forms of hold-out set sampling, with the view to aiding more informed decision making by researchers looking to use this method. Although the ethical considerations are applicable to other health systems, this paper takes a UK focus.

\section{Methods}\label{methods}

\subsection{Setting}\label{setting}

CPMs can exist in a range of settings, using different data sources, and affecting interventions in different ways. We consider the following setting: 

\emph{A CPM is trained on patient-level clinical information (e.g~previous diagnoses) to predict whether a patient will go on to develop an adverse health outcome (e.g.~disease) under standard medical care. Risk scores are then distributed to physicians. They may use these to inform interventions which they recommend to the patient in order to prevent or delay adverse outcomes. Clinicians do not base interventions solely on the risk scores, but also on their expert assessment of the patient (which will generally include information not used by the CPM).}

We assume that patients generally benefit from more accurate CPMs, and that risk scores are interpreted as ``the risk under typical clinical practice if we did not use a CPM to inform interventions''. Furthermore, we assume that a sufficiently accurate CPM provides valuable information to a GP above with which they would otherwise have, allowing them to make an informed decision - this is the cause of performative effects. If drift occurs, the CPM gradually becomes less accurate\cite{Davis2020DetectionUpdating}. To address this, CPMs need to be updated (re-trained) using recent data. If there are performative effects, then the data used to re-train the model will reflect the actions performed in response to predictions made by the original CPM. New CPMs fitted directly to these data will only estimate the risk of the outcome in the setting where the CPM was already in use. This may be a poor approximation of the ``risk under typical practice'' above. For instance, in the EuroSCORE2 score \cite{Nashef2012EuroscoreII}, a risk score fitted directly to population data would approximate an individual's risk of heart attacks after their doctor had already seen and possibly acted on their existing EuroSCORE2 score.

The use of a causal framework to model the effect of CPM-informed interventions has been proposed as a solution this issue\cite{Sperrin2019ExplicitSuccess,Lenert2019PrognosticUnless...}. However, direct measurement or recording of such interventions is often impractical, particularly in most UK healthcare settings which lack the required digital infrastructure. To safely update a CPM in the presence of performative effects, we argue the need to have up-to-date data which reflects typical clinical practice without a CPM. This is tantamount to the use of a hold-out set.

\subsection{Necessity of hold-out sets}\label{necessity}

\begin{figure*}
\centering
\includegraphics[width=\textwidth]{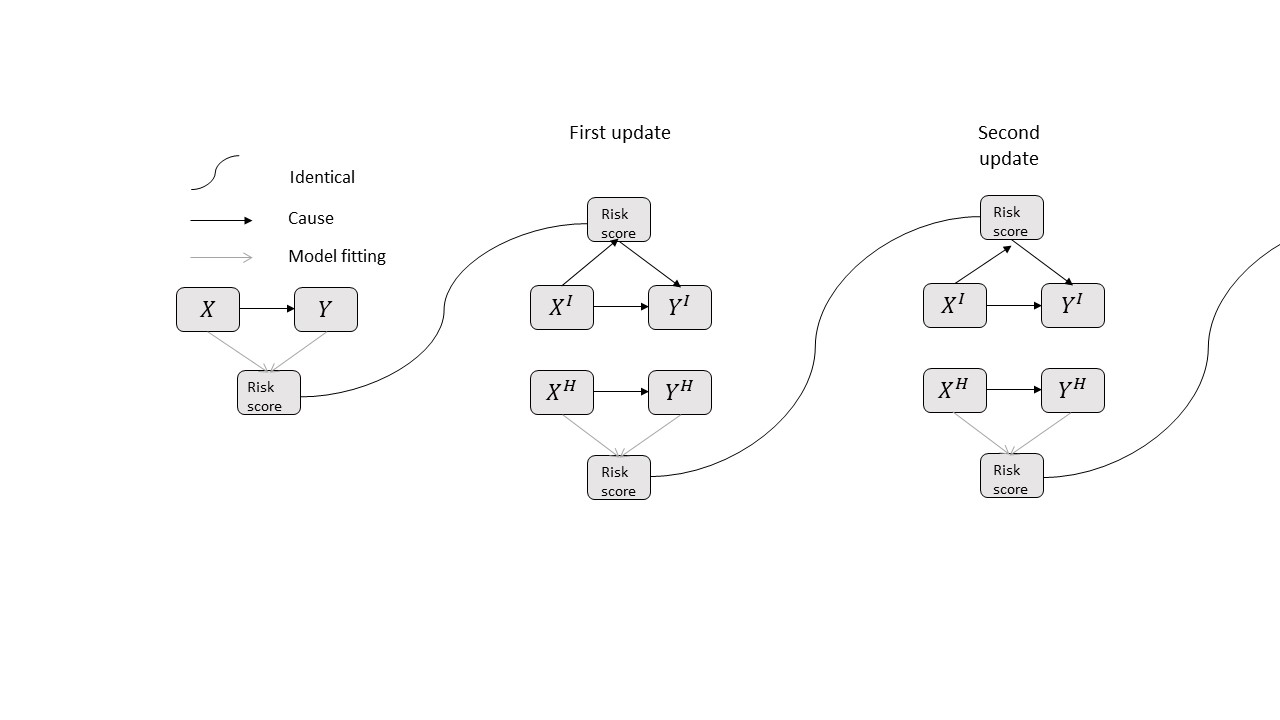}
\caption{Dynamics of a CPM trained once and updated twice using hold-out set methodology. Squares containing $X$ and $Y$ denote covariates and outcome respectively, with superscripts $I$ and $H$ denoting mutually exclusive intervention and hold-out sets.}
\label{fig}
\end{figure*}

The use of hold-out sets modify the setting described above is as follows:

\emph{A CPM is trained and mutually exclusive and complementary ``hold-out'' and ``intervention'' sets are sampled, which we will refer to as $(X^H, Y^H)$ and $(X^I, Y^I)$ respectively, where the superscript refers to hold-out ($H$) or intervention set ($I$). Risk scores are then distributed to physicians, but only for patients in the intervention set. Patients in the hold-out set have access to the same interventions as those in the intervention set, but do not have risk scores distributed to physicians which could inform decision making. When the CPM is updated, it uses data exclusively from the ``hold-out'' set to train the model. Figure \ref{fig} displays the causal dynamics of the system, including how hold-out sets $(X^H,Y^H)$ are used to retrain CPMs, and how performative effects occur in the intervention set $(X^I,Y^I)$.}

Although the use of hold-out sets is not yet standard, and alternative options may be preferable in some cases, we argue toward their general necessity when monitoring and updating CPMs. Other possible methods such as causal modelling approaches have been suggested \cite{Sperrin2019ExplicitSuccess}. We argue that most CPMs will have a performative effect: indeed, CPMs are typically designed in order to bring about an effect on the outcome $Y$. Furthermore, medical systems or care pathways change with time, often deliberately, inducing non-performative drift. This can cause the accuracy of CPMs to deteriorate, resulting in the need to update. Finally, most CPMs are used to guide interventions in complex circumstances, in which it is not possible to exactly specify actions which came about as a result of the CPM (for instance, the degree to which a decision to operate was directly influenced a patient's EuroSCORE II assessment). This precludes direct measurement of the effect of the CPM on interventions and, in turn, on patient outcomes. Hence, this limits the use of causal modelling approaches at solving the problem of performative prediction when updating CPMs. By re-training a CPM on patients whose treatment is not influenced by the risk score, we continue to predict the ``risk of outcome under typical clinical practice''.

Hold-out sets could be used to both monitor and update a CPM. Evaluating the effects of a risk score guided intervention system is possible using a randomly sampled hold-out set, in which the only difference between hold-out and intervention set is the use of CPM derived risk scores, akin to a treatment in clinical trials. Any ``treatment effect'' that comes with use of the model is then easily derived, with the control group being standard medical care. However, our arguments frame the use of hold-out sets primarily as a mechanism to update CPMs in the presence of performative effects, rather than as a research tool to gain knowledge. Additionally, hold-out sets can be used to retrain a CPM when enough drift has occurred to cause a deterioration in accuracy. By retraining on the hold-out set, performative effects of the risk scores are not present in the training data, ensuring risk scores more accurately reflect standard medical care (without a CPM).

\subsection{Sampling hold-out sets}\label{methodology}

We consider three sampling methods for hold-out sets; simple random sampling, cluster randomised sampling and voluntary response sampling\cite{Berndt2020SamplingMethods}. We will explore the merits and drawbacks of each. We note that existing analysis of hold-out sets implicitly consider only simple randomised sampling\cite{Haidar-Wehbe2022OptimalUpdating}. 

In a simple random sampling framework, hold-out sets are drawn as a uniform random sample of the population without the explicit informed consent of patients. This creates a hold-out set with a high degree of external validity. From a statistical perspective, this makes this form of sampling ideal, as it does not introduce bias into the hold-out set\cite{Berndt2020SamplingMethods}.

In cluster randomised sampling, the population is split into clusters (e.g~hospitals or geographic areas), with the hold-out set consisting of patients from a number of randomly selected clusters. Informed consent would not be gained from individuals. If the clusters in the hold-out set are biased in some way, then this will likely worsen the external validity of the hold-out set\cite{Berndt2020SamplingMethods}.

In a voluntary response setting, patients could explicitly consent to be included in a hold-out set. However, this would leave a number of statistical issues. Those who volunteer would likely not constitute a representative sample of the population, meaning the hold-out set would likely represent a biased sample\cite{Berndt2020SamplingMethods}. This means that a model re-trained on the hold-out set would have poor external validity. Furthermore, it may be impractical to seek consent from sufficiently many volunteers to form an adequately sized hold-out set.

Precise comparison of statistical properties of the different types of hold-out set sampling (e.g.~sample size requirements) remains an open problem which is beyond the scope of this work.

\subsection{Ethical considerations}\label{considerations}

We principally consider the fundamental ethical principles of beneficence, non-maleficence, autonomy and justice\cite{Varkey2021PrinciplesPractice}. We will assess hold-out set sampling methods against these ethical principles, alongside some further principles which derive from these.

\subsubsection{Beneficence:}\label{beneficence}

The principle of beneficence requires that treatments or policies benefit the individuals in the target population \cite{Coughlin2008HowEthics}. This presents an apparent dilemma. The use of hold-out sets can lead to a conflict between individual welfare (for those in the hold-out set), and common welfare. On a per-patient basis, if a sufficiently accurate CPM-derived risk score is available, the principal of beneficence obliges the physician to consider the risk score in their decision making process on intervention/treatment\cite{Varkey2021PrinciplesPractice}. However, this may lead to the presence of performative effects which, if not using hold-out sets, could result in a generally worse score for each individual in the population after the CPM is updated. The negative effects of inaccurate CPM-derived risk scores can accrue without bound over the population.  

Generally, patients would have an expectation for physicians to act in their best interests, and thus to universally consider available CPMs, eventually meaning that the CPM becomes inaccurate (and hence we lose benefit for all patients). In simple- and cluster- randomised sampling, given that consent would not be sought, the principle of beneficence may be broken at the individual level. However, collectively (applied to each individual in the total patient population), beneficence indicates that a medical regulator should take the option which results in more accurate risk scores. This suggests use of a randomised (or cluster-randomised) hold-out set, since any other option leads to a less accurate updated CPM.

Cluster randomised sampling ensures an equivalent standard of care for patients in the same cluster, although means there are differing standards between clusters. This would ensure that at a patient level, physicians could still act in the best interests of patients using all available information to them at that time, but it would result in differences between outcomes for populations in different clusters (postcode lotteries of care).

Voluntary response sampling somewhat avoids the per-patient violation of the principle of beneficence, since patients are knowingly agreeing to turn down the benefits associated to the use of a CPM (even with informed consent gained for those in the hold-out set, a physician must still act in the best interest of a patient). However, since this approach will typically lead to a biased hold-out set and hence a less accurate model than would be attained with a randomised hold-out set, use of a voluntary response hold-out set may violate the principle of beneficence on a population scale.

\subsubsection{Non-maleficence:}
\label{nonmalificence}

Considerations of non-maleficence \cite{Summers2009PrinciplesEthics}\cite{Guraya2014EthicsResearch} largely mirror those of beneficence. Gaining informed consent from patients does not give a blanket indemnity to use hold-out sets in scenarios where direct harm could occur to individuals due to inclusion in the hold-out set. It is thus vital that risks to patients are minimised after a decision is made on whether the patient receives a risk score. Depending on the outcome that is predicted by the CPM, patients in the hold-out set may be at risk of harm if they are not intervened on. In those cases, it may not be ethically justifiable to use hold-out sets even with consenting volunteers.

However, there is also non-maleficence considerations when employing CPMs without a hold-out set. If lack of a hold-out set leads to inaccurate risk scores, particularly for risk scores which underestimate risk for high risk patients, then those patients may be at risk of harm due to misdiagnosis or lack of interventions which may have otherwise been applied. 

\subsubsection{Autonomy:}\label{autonomy}

Generally, the use of hold-out sets may be regarded as withholding valuable information with which a patient may make rational, autonomous decisions. That is to say that, although a physician can recommend an intervention, a patient has the right to make the final decision. However, withholding a risk score does not affect the set of interventions available to a patient. Moreover, patients always retain the autonomy to go ahead with any interventions agreed with a physician in any form of hold-out sets.

The principle of autonomy in our setting could cover not only what choice is available to patients from a intervention perspective, but also whether a patient has had the chance to choose whether a CPM is used in their case. In the case of cluster randomised sampling, all patients within a cluster would have the same level of autonomy as each other within that cluster, resulting in an opportunity for a CPM to inform clinical decisions at a cluster level. However, given patients within a cluster randomised hold-out set sampling frame would not have given explicit informed consent to be in the hold-out, or not, dataset, this could be seen to be an unreasonable withdrawal of patient autonomy.

Voluntary response sampling may be able to solve some issues regarding patient autonomy. In particular, patients would be volunteering to have risk scores withheld from themselves and physicians. Provided the patient maintains the ability to withdraw consent, they may still have ultimate autonomy over possible future interventions. Hence, voluntary response sampling generally may provide a greater level of autonomy to patients than randomised sampling without informed consent.

\subsubsection{Justice:}\label{justice}

One key consideration in any hold-out set CPM framework is that of distributive justice - that is, the fair distribution of healthcare services and treatments to patients.

If a CPM is trained on biased data, either on initial training, or when updating using a hold-out set, it may produce a model which disproportionately benefits over-represented groups \cite{Chen2021AlgorithmHealthcare}. Furthermore, it is possible for models to be unfair due to other modelling aspects even when the training data itself is an unbiased sample of the population. We nevertheless argue towards a unbiased hold-out set as necessary for justice considerations.

Simple random sampling has the benefit of not systematically benefiting or harming certain protected classes such as race or gender by over/under inclusion in the hold-out set, relative to presence of these classes in the data source (as long as the general training dataset for the CPM is unbiased\cite{Verheij2018PossibleReuse}).

Justice could be an issue in the case of cluster randomised sampling. There would need to be careful consideration given to whether any classes are being systematically selected due to availability or compromised positions. Given the geographic variance in health outcomes in the UK \cite{Walsh2010ItsOutcomes}, it may also be difficult to obtain a truly representative sample, leading to a biased dataset and ultimately a less accurate CPM.

Voluntary response sampling is likely to lead to biased samples. In particular, it may lead to over-representation of certain groups in the hold-out set who are more likely to volunteer, known as volunteer bias. This occurs in traditional research cohorts such as the UK Biobank \cite{Swanson2012TheBias}. Volunteer bias will impact new risk scores upon model retraining, and may lead to worse accuracy amongst under-represented groups. Additionally, if certain groups volunteer for the hold-out set at greater rates, these groups will be disproportionately affected by any potential harms that presence in a hold-out set may bring. As discussed above, non-minimal harms to patients may not be ethically viable even with informed consent gained from patients.

Issues concerning incentives for volunteering arise when considering the principle of justice. Researchers and physicians must take care to not unduly pressure patients to volunteer to be in the hold-out set, as this could lead to systematically over-represented classes of ``pressured'' patients due to compromised position or availability. Correspondingly, if a group is known to be under-represented in a voluntary hold-out set, this may lead to patients volunteering in order to ensure representation of their own groups, which may be considered an unreasonable incentive.

\subsubsection{Informed consent:}\label{consent}

In the UK, it is not necessarily legally required to obtain informed consent from patients whose electronic health records data are used to train a CPM, even in the absence of any use of a hold-out set\cite{Taylor2016DirectStatutes}. Patients can, however, opt out of having their data used for research\cite{NHS2022ProtectingData}. Furthermore, a patient does not need to give consent for a CPM to be used to aid decision making.

However, in a hold-out setting, there is an additional consideration. By withholding risk scores from certain individuals, we potentially deny them use of tools which could positively impact their outcomes. 
Withholding risk scores is not necessarily in the interests of the individual patient in the hold-out set; rather, it is in the interests of the population as a whole and potentially for the benefit of research. From an ethical perspective, informed consent may thus be necessary to withhold the use of risk scores in this way.

To guarantee an unbiased source of data which can be used to safely retrain prediction models which generalise to the full population, hold-out sets need to be randomly sampled. Seeking consent from those in the hold-out set would be highly likely to cause participation bias\cite{Berndt2020SamplingMethods}, and logistically within the NHS infrastructure it may not be feasible. However, if the goal is a more generalisable model, absence of informed consent is necessary. In any CPM-intervention setting, use of hold-out sets must be weighed against any risks to patients which arise as a result of not using a CPM. Therefore, the use of hold-out sets without informed consent must be weighed against the principle of non-maleficence as a priority, and is highly setting dependent. This argument extends to both simple and cluster randomised sampling.

In the case of voluntary response sampling, informed consent would be sought from all those in the hold-out set, but not necessarily those in the intervention set, which is seen as the default. This means that, ideally, patients in the hold-out set would be fully aware of any potential risks or lost benefits resulting from the withholding of risk scores. It should be noted that these patients would have access to the same interventions as those in the intervention set. However, risks to patients in the hold-out set would still need to be minimised, and in some settings this may not be acceptable even with consent. Furthermore, bias from voluntary response sampling may increase risks to individuals in the patient population through the use of less accurate risk scores.

\subsubsection{Clinical equipoise:}\label{equipoise}

The assumption of clinical equipoise is a key component of the ethical argument for randomisation in randomised control trials\cite{Cook2011ClinicalTrials}. Our setting is notably different in that use of an accurate CPM is assumed to improve the expected outcome for a patient, usually due to improved prognosis of a disease or allocation of treatments.
A hold-out set would be used in one of two scenarios: either risk score informed interventions have been proven to be more effective than otherwise, in which case there is not clinical equipoise; or risk score informed interventions perform no better than interventions without a risk score, in which case there is no longer an argument for use of a CPM in the first place. If clinical equipoise is assumed, this would only be the case on initial deployment of the CPM, until its effectiveness has been established.

The choice of sampling, including whether or not consent is gained, does not necessarily affect these arguments. The argument for use of a hold-out set relies on the intent that it will maintain an optimal allocation of interventions, rather than as a research tool to gain knowledge. This differs from randomised control trials, in which the primary purpose is to gain knowledge, and the medical community is assumed to be in equipoise over the effect of the potential treatment. A clinical trial may or may not improve the healthcare system over a steady baseline defined by existing treatment, whereas the corresponding baseline in the CPM updating setting deteriorates due to drift, and we seek to restore the baseline (but are not in equipoise over whether updating the CPM will do so).

\subsubsection{Challenges to shared decision-making:}\label{truth}

One key principle in modern medical ethics is that of trust and shared decision-making between patient and clinician. One aspect of this is patient autonomy, covered above \cite{Gillon2015DefendingEthics}. However, in order to achieve this ideal of trust and shared-decision making, it is also necessary that a clinician is truthful with their patient.

When considering the principle of truth telling, it must be noted that how much information a doctor divulges to a patient may differ across cultures\cite{Tuckett2004Truth-tellingLiterature}. Furthermore, it must be balanced with the principles of beneficence and patient autonomy\cite{Sullivan2001Truth-tellingDiagnoses}. Nevertheless, the principle of truth telling  is potentially violated by the use of hold-out sets by design, as risk scores are not generated for some patients and therefore the score is withheld from the clinical-patient discussion. 

Once again, evaluation of this principle is situation dependent. It should be noted that using simple random sampling in the motivating setting, the physician would still be able to make a decision on the information present, or to evaluate risk of a future outcome without the use of the risk score for the hold-out set patients. Therefore this information may not necessarily be vital to the patient's outcome. However, a patient who is unaware that they are in the hold-out set may reasonably expect to be given a risk score, if they were aware of what this means and that it were possible to generate one for them.

Cluster randomised hold-out sets violate the principal of truth telling in the same way as simple random sampling. Whilst one could argue that patients in the hold-out set who do not receive risk scores would now be receiving the standard care of their cluster, a patient would still reasonably expect that if it were possible to generate a risk score for them, then they would want a physician to make use of this information.

In a voluntary response setting, providing a patient with the ability to withdraw consent at any time would ensure that a patient could always have access to a risk score upon request.

\subsection{Case studies}\label{casestudies}

In this section, some potential situation specific ethical issues are discussed through the use of two real-world models: the Scottish Patients at Risk of Readmission and Admission (SPARRA) model\cite{Liley2023DevelopmentScotland} and the Epic Sepsis Model (ESM)\cite{Wong2021ExternalPatients}.

\textbf{SPARRA:} \emph{This is a model to predict the 1-year risk of emergency admission to hospital based on electronic health records in Scotland. SPARRA risk scores are calculated monthly, and individual-level scores can be accessed by GPs. GPs may choose to act on these scores with high-risk patients to lower their risk of emergency admission, such as by choosing to give an enhanced follow-up in comparison to lower-risk patients. There are no specific interventions that physicians must take in response to risk scores, as the nature of each patient's risk is unique. It has been demonstrated that preventative interventions influenced by the model have the potential to alter future risk scores, resulting in a potential underestimation of risk in high risk patients\cite{Liley2023DevelopmentScotland}\cite{Liley2021ModelBias}.}

\textbf{ESM:} \emph{The Epic Sepsis Model\cite{Wong2021ExternalPatients} is a prediction model which uses electronic health records to predict the risk of sepsis for patients every 15 minutes. It generates automatic alerts to warn clinicians that a patient may be becoming septic, at which point appropriate interventions can be taken. In the ICU setting, we would expect that interventions may occur in under 15 minutes, and such interventions may be life-saving; hence, a patient for whom the ESM score leads to an intervention may be observed to have a lower risk of sepsis than predicted by a well-calibrated risk score, leading to performative effects in an updated risk score. }

In the case of the SPARRA model, the outcome being predicted is not imminently affecting the patient. Emergency admission could also be caused by a wide variety of factors, and interventions that may be applied are highly patient dependent. A physician and patient are likely to make decisions about interventions based on a wide variety of factors, of which the model predicted risk score is only a small factor. In this case, withholding risk scores from patients is less likely to break the principle of non-maleficence, as any harms due to withholding the risk score are likely to be non-immediate and minimal. Furthermore, the use of hold-out sets generally in this setting would provide additional value to the full patient population by way of more accurate pre-intervention risk scores.

For the ESM one may make the argument in favour of a hold-out set in this scenario, specifically that the harms to patients as a result of inaccurate risk scores would be greater in this setting, due to the nature of the outcome being predicted. However, we argue that this is not enough of a reason to use hold-out sets. In the case of the ESM, the outcome being predicted poses an immediate risk to patients. Furthermore, it is a model used on patients who are already hospitalised, and therefore highly vulnerable. Gaining consent to be included in a hold-out set from such patients would clearly be unethical, as they are in a compromised state. Furthermore, under (or over) estimation of risk presents a much greater risk to patients in the case of ESM in comparison to SPARRA. Withholding risk scores from some patients for the ESM may increase the risk of misdiagnosis, particularly false negatives. This presents severe possible harms to patients in the hold-out set, and breaks the principle of non-maleficence.  This can then be viewed as a scenario where the use of hold-out sets, with or without consent, would be a gross violation of widely accepted clinical research ethics.  

The two models have been chosen specifically due to the stark contrast in the nature of the predicted outcomes, as well as the risk profiles of patients. This highlights the wide variety of different CPMs that exist in the real world. Any ethical questions which arise due to the potential use of hold-out sets must be contextualised to the model in question, and treated uniquely.

\subsection{Hold-out sets to measure the effectiveness of a CPM}

So far in this paper we have mainly considered hold-out sets as a tool for updating CPMs, and subsequent ethical arguments have followed from that. There is however, the possibility of using hold-out sets to measure the effectiveness of a CPM compared to standard medical care.

Typically, CPMs are validated in-silico, for example using discrimination or calibration metrics and cross-validation \cite{Staffa2021StatisticalModels}. However, this fails to capture the overall effectiveness of a CPM. Indeed, the latter depends on a variety of factors that go beyond the accuracy of the predictions. Among others, this includes: how well is the CPM integrated into the healthcare system (can clinicians or patient easily access the scores?), how well is the CPM integrated into the decision-making process (if a CPM is available to them, do clinicians and patients take it into account when deciding the best course of action?) and the effectiveness of the possible interventions themselves (does the intervention actually reduce the risk of an adverse event?). 

If a hold-out set is a random sample from the population, then generally you may be able to say that any difference in downstream case numbers between the hold-out and intervention set is due to the use of the CPM. This allows for analysis to be done in a similar way to a randomised control trial. For example, one study utilised a randomised stepped wedge trial design to assess the effects (both in terms of patient outcomes and resource utilisation) associated to the introduction of a CPM (to predict the risk of emergency hospital admissions) into the Welsh primary care system \cite{Snooks2019EffectsTrial}. It should be noted that simple random sampling would then offer the most appropriate sampling method for this use-case. Likewise, cluster randomised sampling would have potential to also offer unbiased results, provided that clusters were chosen appropriately. However, voluntary response sampling, as previously discussed, would lead to biased samples. If the intent is to measure the overall effectiveness of a CPM, voluntary response would make this very difficult to interpret.

\subsection{Summary and Recommendations}\label{recommendations}

In general we expect that the greater the risk to the patient from having no risk score (and hence the more unappealing the hold-out set approach appears), the worse the cost from having an inaccurate risk score (so the greater the incentive for a randomised hold-out set).

The necessity of use of a hold-out set at all is dependent on drift and performative effects. The presence of drift is largely independent of the severity of the outcome predicted by the CPM. Unfortunately, in settings where ignoring a CPM has a severe consequence, performative effects are necessarily present. Recommendations can not be made at a general level, as it depends on the severity of such consequences. However, considerations of the environment in which a CPM operates can help inform choice of sampling method for a hold-out set, or whether to use a hold-out set at all. In particular, the nature of the outcome being predicted by the CPM is an important factor. If the outcome being predicted is one which will not cause immediate serious harm to patients, and the patients themselves are not in a compromised position, then a setting specific argument can be made for use of a hold-out set without consent. Use of cluster randomisation for hold-out sets may offer additional ethical safeguards for patients without sacrificing statistical validity in terms of bias, provided that appropriate clusters are chosen. Specifically, cluster randomisation ensures that patients always receive the same care as others in their cluster. However, care must be taken to select clusters which do not over-sample certain groups, otherwise issues of justice and fairness arise.

Voluntary response hold-out sets, whilst avoiding some of the ethical issues discussed in this paper, offer the potential to indirectly harm patients on the whole through inaccurate risk scores trained on a biased hold-out set. It may be possible to ensure that a voluntary response hold-out set is balanced across some protected classes or important characteristics. However, this does not correct for different rates of volunteering across unobserved patient characteristics.

Other approaches have been suggested for mitigating against the effects of performative prediction in the presence of external drift. In particular, causal modelling approaches have been suggested \cite{Sperrin2019ExplicitSuccess}. Whilst this may be possible, it relies on a more mature data collection system than currently exists in any area of the health care system, in particular the recording of all interventions, and the collection of all variables which exist in the causal structure. Thus, in the absence of this, hold-out sets may be the only viable method of updating risk scores in the presence of performative effects.

Any implementation of hold-out sets will need to take into account the views of key stakeholders, clinicians and patients themselves. In particular, use of hold-out sets without informed consent may risk damaging trust in the health system and lead to unintended consequences, therefore much more research is needed.

\section{Discussion}\label{discussion}

Our work principally suggests that the ethical viability of hold-out sets is setting dependent. Furthermore, certain implementations of hold-out sets may be more appropriate than others. The key advantage of hold-out sets is that they offer the possibility to distribute more accurate pre-intervention risk scores to those in the intervention set. On an aggregate level this may be more optimal than distributing risk scores to all patients, given some objective function to be optimised such as lowering the number of patients with $Y=1$ over the life-span of the CPM. However from an ethical perspective, this is may not be enough of a reason to use hold-out sets, particularly without informed consent. Withholding risk scores must not cause an unacceptable risk of harm to patients, and must respect principles of patient autonomy and justice. Ultimately, any implementation of hold-out sets in a health setting in the absence of informed consent must respect patients, putting their health and well-being first. Cumulative benefits to the group must be balanced against individual considerations at a patient level, including risk of harm, lack of autonomy or potential justice issues.

It is critical to realise that a choice must be made between not updating a CPM, updating a CPM without a hold-out set, and updating with a hold-out set. All have drawbacks at an individual or population level. Each option violates one or more ethical principles, as is typically the case in research ethics, and violation of principles thus cannot preclude use of a method: risks must be weighed against each other. We do note however that in the absence of recorded interventions in response to risk scores and therefore explicit causal modelling, hold-out sets may be the only viable way of updating in the presence of performative effects.

This paper does not seek to prescribe hold-out sets as a catch-all solution to performative prediction in CPMs, nor which implementations of sampling are universally ethically viable. Rather, it seeks to provide an initial discussion of some of the general issues in this field in the absence of existing literature. In any possible implementation of hold-out sets, a sampling framework should be implemented which offers the lowest possible degree of bias in the hold-out set given that ethical issues have been considered. This paper further highlights the need for robust study protocols and wide involvement from scientists from a range of disciplines, health practitioners and patients themselves. Furthermore, this paper does not seek to analyse the potential for hold-out sets to be practical within the NHS (or other health systems) infrastructure, although this will be a necessary consideration. Without these considerations, hold-out sets as a tool have the potential to lead to gross violations of widely accepted research ethics principles. Context dependent ethical discussion is necessary before use of a hold-out set in a CPM can be considered.

\section{Conflicts of Interest}
The Author(s) declare(s) that there is no conflict of interest

\section{Funding}
Louis Chislett acknowledges the receipt of studentship awards from the Health Data Research UK-The Alan Turing Institute Wellcome PhD Programme in Health Data Science (Grant Ref: 218529/Z/19/Z)

This work was supported by EPSRC Grant EP/X03870X/1 \&\ The Alan Turing Institute’

\bibliography{references}


\begin{thebibliography}{34}
\ifx \bisbn   \undefined \def \bisbn  #1{ISBN #1}\fi
\ifx \binits  \undefined \def \binits#1{#1}\fi
\ifx \bauthor  \undefined \def \bauthor#1{#1}\fi
\ifx \batitle  \undefined \def \batitle#1{#1}\fi
\ifx \bjtitle  \undefined \def \bjtitle#1{#1}\fi
\ifx \bvolume  \undefined \def \bvolume#1{\textbf{#1}}\fi
\ifx \byear  \undefined \def \byear#1{#1}\fi
\ifx \bissue  \undefined \def \bissue#1{#1}\fi
\ifx \bfpage  \undefined \def \bfpage#1{#1}\fi
\ifx \blpage  \undefined \def \blpage #1{#1}\fi
\ifx \burl  \undefined \def \burl#1{\textsf{#1}}\fi
\ifx \doiurl  \undefined \def \doiurl#1{\url{https://doi.org/#1}}\fi
\ifx \betal  \undefined \def \betal{\textit{et al.}}\fi
\ifx \binstitute  \undefined \def \binstitute#1{#1}\fi
\ifx \binstitutionaled  \undefined \def \binstitutionaled#1{#1}\fi
\ifx \bctitle  \undefined \def \bctitle#1{#1}\fi
\ifx \beditor  \undefined \def \beditor#1{#1}\fi
\ifx \bpublisher  \undefined \def \bpublisher#1{#1}\fi
\ifx \bbtitle  \undefined \def \bbtitle#1{#1}\fi
\ifx \bedition  \undefined \def \bedition#1{#1}\fi
\ifx \bseriesno  \undefined \def \bseriesno#1{#1}\fi
\ifx \blocation  \undefined \def \blocation#1{#1}\fi
\ifx \bsertitle  \undefined \def \bsertitle#1{#1}\fi
\ifx \bsnm \undefined \def \bsnm#1{#1}\fi
\ifx \bsuffix \undefined \def \bsuffix#1{#1}\fi
\ifx \bparticle \undefined \def \bparticle#1{#1}\fi
\ifx \barticle \undefined \def \barticle#1{#1}\fi
\bibcommenthead
\ifx \bconfdate \undefined \def \bconfdate #1{#1}\fi
\ifx \botherref \undefined \def \botherref #1{#1}\fi
\ifx \url \undefined \def \url#1{\textsf{#1}}\fi
\ifx \bchapter \undefined \def \bchapter#1{#1}\fi
\ifx \bbook \undefined \def \bbook#1{#1}\fi
\ifx \bcomment \undefined \def \bcomment#1{#1}\fi
\ifx \oauthor \undefined \def \oauthor#1{#1}\fi
\ifx \citeauthoryear \undefined \def \citeauthoryear#1{#1}\fi
\ifx \endbibitem  \undefined \def \endbibitem {}\fi
\ifx \bconflocation  \undefined \def \bconflocation#1{#1}\fi
\ifx \arxivurl  \undefined \def \arxivurl#1{\textsf{#1}}\fi
\csname PreBibitemsHook\endcsname

\bibitem[\protect\citeauthoryear{Hastie et~al.}{2001}]{Hastie2001TheLearning}
\begin{bbook}
\bauthor{\bsnm{Hastie}, \binits{T.}},
\bauthor{\bsnm{Friedman}, \binits{J.}},
\bauthor{\bsnm{Tibshirani}, \binits{R.}}:
\bbtitle{{The Elements of Statistical Learning}}.
\bpublisher{Springer},
\blocation{New York, NY}
(\byear{2001}).
\doiurl{10.1007/978-0-387-21606-5} .
\burl{https://link.springer.com/book/10.1007/978-0-387-21606-5}
\end{bbook}
\endbibitem

\bibitem[\protect\citeauthoryear{Topol}{2019}]{Topol2019High-performanceIntelligence}
\begin{botherref}
\oauthor{\bsnm{Topol}, \binits{E.J.}}:
{High-performance medicine: the convergence of human and artificial intelligence}.
Nature Publishing Group
(2019).
\doiurl{10.1038/s41591-018-0300-7}
\end{botherref}
\endbibitem

\bibitem[\protect\citeauthoryear{Cowley et~al.}{2019}]{Cowley2019MethodologicalLiterature}
\begin{botherref}
\oauthor{\bsnm{Cowley}, \binits{L.E.}},
\oauthor{\bsnm{Farewell}, \binits{D.M.}},
\oauthor{\bsnm{Maguire}, \binits{S.}},
\oauthor{\bsnm{Kemp}, \binits{A.M.}}:
{Methodological standards for the development and evaluation of clinical prediction rules: a review of the literature}.
Diagnostic and Prognostic Research
\textbf{3}(1)
(2019)
\doiurl{10.1186/s41512-019-0060-y}
\end{botherref}
\endbibitem

\bibitem[\protect\citeauthoryear{Nashef et~al.}{2012}]{Nashef2012EuroscoreII}
\begin{barticle}
\bauthor{\bsnm{Nashef}, \binits{S.A.M.}},
\bauthor{\bsnm{Roques}, \binits{F.}},
\bauthor{\bsnm{Sharples}, \binits{L.D.}},
\bauthor{\bsnm{Nilsson}, \binits{J.}},
\bauthor{\bsnm{Smith}, \binits{C.}},
\bauthor{\bsnm{Goldstone}, \binits{A.R.}},
\bauthor{\bsnm{Lockowandt}, \binits{U.}}:
\batitle{{Euroscore II}}.
\bjtitle{European Journal of Cardio-thoracic Surgery}
\bvolume{41}(\bissue{4}),
\bfpage{734}--\blpage{745}
(\byear{2012})
\doiurl{10.1093/ejcts/ezs043}
\end{barticle}
\endbibitem

\bibitem[\protect\citeauthoryear{{\v{Z}}liobait{\.{e}}}{2010}]{Zliobaite2010LearningOverview}
\begin{barticle}
\bauthor{\bsnm{{\v{Z}}liobait{\.{e}}}, \binits{I.}}:
\batitle{{Learning under Concept Drift: an Overview}}.
\bjtitle{arXiv preprint}
(\byear{2010})
\doiurl{10.48550/arXiv.1010.4784}
\end{barticle}
\endbibitem

\bibitem[\protect\citeauthoryear{Davis et~al.}{2020}]{Davis2020DetectionUpdating}
\begin{botherref}
\oauthor{\bsnm{Davis}, \binits{S.E.}},
\oauthor{\bsnm{Greevy}, \binits{R.A.}},
\oauthor{\bsnm{Lasko}, \binits{T.A.}},
\oauthor{\bsnm{Walsh}, \binits{C.G.}},
\oauthor{\bsnm{Matheny}, \binits{M.E.}}:
{Detection of calibration drift in clinical prediction models to inform model updating}.
Journal of Biomedical Informatics
\textbf{112}
(2020)
\doiurl{10.1016/j.jbi.2020.103611}
\end{botherref}
\endbibitem

\bibitem[\protect\citeauthoryear{Finlayson et~al.}{2021}]{Finlayson2021TheIntelligence}
\begin{barticle}
\bauthor{\bsnm{Finlayson}, \binits{S.G.}},
\bauthor{\bsnm{Subbaswamy}, \binits{A.}},
\bauthor{\bsnm{Singh}, \binits{K.}},
\bauthor{\bsnm{Bowers}, \binits{J.}},
\bauthor{\bsnm{Kupke}, \binits{A.}},
\bauthor{\bsnm{Zittrain}, \binits{J.}},
\bauthor{\bsnm{Kohane}, \binits{I.S.}},
\bauthor{\bsnm{Saria}, \binits{S.}}:
\batitle{{The Clinician and Dataset Shift in Artificial Intelligence}}.
\bjtitle{New England Journal of Medicine}
\bvolume{385}(\bissue{3}),
\bfpage{283}--\blpage{286}
(\byear{2021})
\doiurl{10.1056/nejmc2104626}
\end{barticle}
\endbibitem

\bibitem[\protect\citeauthoryear{Perdomo et~al.}{2020}]{Perdomo2020PerformativePrediction}
\begin{bchapter}
\bauthor{\bsnm{Perdomo}, \binits{J.C.}},
\bauthor{\bsnm{Zrnic}, \binits{T.}},
\bauthor{\bsnm{Mendler-D~¨}, \binits{C.}},
\bauthor{\bsnm{Hardt}, \binits{M.}}:
\bctitle{{Performative Prediction}}.
In: \bbtitle{International Conference on Machine Learning}
(\byear{2020})
\end{bchapter}
\endbibitem

\bibitem[\protect\citeauthoryear{Toll et~al.}{2008}]{Toll2008ValidationReview}
\begin{botherref}
\oauthor{\bsnm{Toll}, \binits{D.B.}},
\oauthor{\bsnm{Janssen}, \binits{K.J.M.}},
\oauthor{\bsnm{Vergouwe}, \binits{Y.}},
\oauthor{\bsnm{Moons}, \binits{K.G.M.}}:
{Validation, updating and impact of clinical prediction rules: A review}
(2008).
\doiurl{10.1016/j.jclinepi.2008.04.008}
\end{botherref}
\endbibitem

\bibitem[\protect\citeauthoryear{Sutton et~al.}{2020}]{Sutton2020AnSuccess}
\begin{botherref}
\oauthor{\bsnm{Sutton}, \binits{R.T.}},
\oauthor{\bsnm{Pincock}, \binits{D.}},
\oauthor{\bsnm{Baumgart}, \binits{D.C.}},
\oauthor{\bsnm{Sadowski}, \binits{D.C.}},
\oauthor{\bsnm{Fedorak}, \binits{R.N.}},
\oauthor{\bsnm{Kroeker}, \binits{K.I.}}:
{An overview of clinical decision support systems: benefits, risks, and strategies for success}.
Nature Research
(2020).
\doiurl{10.1038/s41746-020-0221-y}
\end{botherref}
\endbibitem

\bibitem[\protect\citeauthoryear{Hippisley-Cox et~al.}{2017}]{Hippisley-Cox2017DevelopmentStudy}
\begin{botherref}
\oauthor{\bsnm{Hippisley-Cox}, \binits{J.}},
\oauthor{\bsnm{Coupland}, \binits{C.}},
\oauthor{\bsnm{Brindle}, \binits{P.}}:
{Development and validation of QRISK3 risk prediction algorithms to estimate future risk of cardiovascular disease: Prospective cohort study}.
BMJ (Online)
\textbf{357}
(2017)
\doiurl{10.1136/bmj.j2099}
\end{botherref}
\endbibitem

\bibitem[\protect\citeauthoryear{Liley et~al.}{2021}]{Liley2021ModelBias}
\begin{bchapter}
\bauthor{\bsnm{Liley}, \binits{J.}},
\bauthor{\bsnm{Emerson}, \binits{S.R.}},
\bauthor{\bsnm{Mateen}, \binits{B.A.}},
\bauthor{\bsnm{Vallejos}, \binits{C.A.}},
\bauthor{\bsnm{Louis}, \binits{J.M.} \bsuffix{Aslett}},
\bauthor{\bsnm{Vollmer}, \binits{S.J.}}:
\bctitle{{Model updating after interventions paradoxically introduces bias}}.
In: \bbtitle{International Conference on Artificial Intelligence and Statistics},
vol. \bseriesno{130}
(\byear{2021}).
\burl{https://www.who.int/news-room/}
\end{bchapter}
\endbibitem

\bibitem[\protect\citeauthoryear{Lenert et~al.}{2019}]{Lenert2019PrognosticUnless...}
\begin{botherref}
\oauthor{\bsnm{Lenert}, \binits{M.C.}},
\oauthor{\bsnm{Matheny}, \binits{M.E.}},
\oauthor{\bsnm{Walsh}, \binits{C.G.}}:
{Prognostic models will be victims of their own success, unless...}
Oxford University Press
(2019).
\doiurl{10.1093/jamia/ocz145}
\end{botherref}
\endbibitem

\bibitem[\protect\citeauthoryear{Sperrin et~al.}{2019}]{Sperrin2019ExplicitSuccess}
\begin{botherref}
\oauthor{\bsnm{Sperrin}, \binits{M.}},
\oauthor{\bsnm{Jenkins}, \binits{D.}},
\oauthor{\bsnm{Martin}, \binits{G.P.}},
\oauthor{\bsnm{Peek}, \binits{N.}}:
{Explicit causal reasoning is needed to prevent prognostic models being victims of their own success}.
Oxford University Press
(2019).
\doiurl{10.1093/jamia/ocz197}
\end{botherref}
\endbibitem

\bibitem[\protect\citeauthoryear{Berndt}{2020}]{Berndt2020SamplingMethods}
\begin{barticle}
\bauthor{\bsnm{Berndt}, \binits{A.E.}}:
\batitle{{Sampling Methods}}.
\bjtitle{Journal of Human Lactation}
\bvolume{36}(\bissue{2}),
\bfpage{224}--\blpage{226}
(\byear{2020})
\doiurl{10.1177/0890334420906850}
\end{barticle}
\endbibitem

\bibitem[\protect\citeauthoryear{Haidar-Wehbe et~al.}{2022}]{Haidar-Wehbe2022OptimalUpdating}
\begin{barticle}
\bauthor{\bsnm{Haidar-Wehbe}, \binits{S.}},
\bauthor{\bsnm{Emerson}, \binits{S.R.}},
\bauthor{\bsnm{Aslett}, \binits{L.J.M.}},
\bauthor{\bsnm{Liley}, \binits{J.}}:
\batitle{{Optimal sizing of a holdout set for safe predictive model updating}}.
\bjtitle{arXiv preprint}
(\byear{2022})
\doiurl{10.48550/arXiv.2202.06374}
\end{barticle}
\endbibitem

\bibitem[\protect\citeauthoryear{Varkey}{2021}]{Varkey2021PrinciplesPractice}
\begin{botherref}
\oauthor{\bsnm{Varkey}, \binits{B.}}:
{Principles of Clinical Ethics and Their Application to Practice}.
S. Karger AG
(2021).
\doiurl{10.1159/000509119}
\end{botherref}
\endbibitem

\bibitem[\protect\citeauthoryear{Coughlin}{2008}]{Coughlin2008HowEthics}
\begin{barticle}
\bauthor{\bsnm{Coughlin}, \binits{S.S.}}:
\batitle{{How Many Principles for Public Health Ethics?}}
\bjtitle{The Open Public Health Journal}
\bvolume{1}(\bissue{1}),
\bfpage{8}--\blpage{16}
(\byear{2008})
\doiurl{10.2174/1874944500801010008}
\end{barticle}
\endbibitem

\bibitem[\protect\citeauthoryear{Summers and Morrison}{2009}]{Summers2009PrinciplesEthics}
\begin{bchapter}
\bauthor{\bsnm{Summers}, \binits{J.}},
\bauthor{\bsnm{Morrison}, \binits{E.}}:
\bctitle{{Principles of Healthcare Ethics}}.
In: \bbtitle{Health Care Ethics},
\bedition{2}nd edn.,
pp. \bfpage{41}--\blpage{58}.
\bpublisher{Jones and Bartlett Publishers}, \blocation{???}
(\byear{2009})
\end{bchapter}
\endbibitem

\bibitem[\protect\citeauthoryear{Guraya et~al.}{2014}]{Guraya2014EthicsResearch}
\begin{barticle}
\bauthor{\bsnm{Guraya}, \binits{S.Y.}},
\bauthor{\bsnm{London}, \binits{N.J.M.}},
\bauthor{\bsnm{Guraya}, \binits{S.S.}}:
\batitle{{Ethics in medical research}}.
\bjtitle{Journal of Microscopy and Ultrastructure}
\bvolume{2}(\bissue{3}),
\bfpage{121}
(\byear{2014})
\doiurl{10.1016/j.jmau.2014.03.003}
\end{barticle}
\endbibitem

\bibitem[\protect\citeauthoryear{Chen et~al.}{2021}]{Chen2021AlgorithmHealthcare}
\begin{barticle}
\bauthor{\bsnm{Chen}, \binits{R.J.}},
\bauthor{\bsnm{Chen}, \binits{T.Y.}},
\bauthor{\bsnm{Lipkova}, \binits{J.}},
\bauthor{\bsnm{Wang}, \binits{J.J.}},
\bauthor{\bsnm{Williamson}, \binits{D.F.K.}},
\bauthor{\bsnm{Lu}, \binits{M.Y.}},
\bauthor{\bsnm{Sahai}, \binits{S.}},
\bauthor{\bsnm{Mahmood}, \binits{F.}}:
\batitle{{Algorithm Fairness in AI for Medicine and Healthcare}}.
\bjtitle{arXiv preprint}
(\byear{2021})
\doiurl{10.48550/arXiv.2110.00603}
\end{barticle}
\endbibitem

\bibitem[\protect\citeauthoryear{Verheij et~al.}{2018}]{Verheij2018PossibleReuse}
\begin{botherref}
\oauthor{\bsnm{Verheij}, \binits{R.A.}},
\oauthor{\bsnm{Curcin}, \binits{V.}},
\oauthor{\bsnm{Delaney}, \binits{B.C.}},
\oauthor{\bsnm{McGilchrist}, \binits{M.M.}}:
{Possible sources of bias in primary care electronic health record data use and reuse}.
Journal of Medical Internet Research
\textbf{20}(5)
(2018)
\doiurl{10.2196/JMIR.9134}
\end{botherref}
\endbibitem

\bibitem[\protect\citeauthoryear{Walsh et~al.}{2010}]{Walsh2010ItsOutcomes}
\begin{barticle}
\bauthor{\bsnm{Walsh}, \binits{D.}},
\bauthor{\bsnm{Bendel}, \binits{N.}},
\bauthor{\bsnm{Jones}, \binits{R.}},
\bauthor{\bsnm{Hanlon}, \binits{P.}}:
\batitle{{It's not 'just deprivation': Why do equally deprived UK cities experience different health outcomes?}}
\bjtitle{Public Health}
\bvolume{124}(\bissue{9}),
\bfpage{487}--\blpage{495}
(\byear{2010})
\doiurl{10.1016/j.puhe.2010.02.006}
\end{barticle}
\endbibitem

\bibitem[\protect\citeauthoryear{Swanson}{2012}]{Swanson2012TheBias}
\begin{botherref}
\oauthor{\bsnm{Swanson}, \binits{J.M.}}:
{The UK Biobank and selection bias}.
Elsevier B.V.
(2012).
\doiurl{10.1016/S0140-6736(12)61179-9}
\end{botherref}
\endbibitem

\bibitem[\protect\citeauthoryear{Taylor et~al.}{2016}]{Taylor2016DirectStatutes}
\begin{botherref}
\oauthor{\bsnm{Taylor}, \binits{R.M.}},
\oauthor{\bsnm{Fern}, \binits{L.A.}},
\oauthor{\bsnm{Aslam}, \binits{N.}},
\oauthor{\bsnm{Whelan}, \binits{J.S.}}:
{Direct access to potential research participants for a cohort study using a confidentiality waiver included in UK National Health Service legal statutes}.
BMJ Open
\textbf{6}(8)
(2016)
\doiurl{10.1136/bmjopen-2016-011847}
\end{botherref}
\endbibitem

\bibitem[\protect\citeauthoryear{{NHS}}{2022}]{NHS2022ProtectingData}
\begin{botherref}
\oauthor{\bsnm{{NHS}}}:
{Protecting patient data}
(2022).
\url{https://digital.nhs.uk/services/national-data-opt-out/understanding-the-national-data-opt-out/protecting-patient-data}
\end{botherref}
\endbibitem

\bibitem[\protect\citeauthoryear{Cook and Sheets}{2011}]{Cook2011ClinicalTrials}
\begin{botherref}
\oauthor{\bsnm{Cook}, \binits{C.}},
\oauthor{\bsnm{Sheets}, \binits{C.}}:
{Clinical equipoise and personal equipoise: Two necessary ingredients for reducing bias in manual therapy trials}
(2011).
\doiurl{10.1179/106698111X12899036752014}
\end{botherref}
\endbibitem

\bibitem[\protect\citeauthoryear{Gillon}{2015}]{Gillon2015DefendingEthics}
\begin{botherref}
\oauthor{\bsnm{Gillon}, \binits{R.}}:
{Defending the four principles approach as a good basis for good medical practice and therefore for good medical ethics}.
Technical Report~1
(2015).
\doiurl{10.1136/medethics-2014-102282} .
\url{https://www.jstor.org/stable/43283239}
\end{botherref}
\endbibitem

\bibitem[\protect\citeauthoryear{Tuckett}{2004}]{Tuckett2004Truth-tellingLiterature}
\begin{botherref}
\oauthor{\bsnm{Tuckett}, \binits{A.G.}}:
{Truth-telling in clinical practice and the arguments for and against: A review of the literature}
(2004).
\doiurl{10.1191/0969733004ne728oa}
\end{botherref}
\endbibitem

\bibitem[\protect\citeauthoryear{Sullivan et~al.}{2001}]{Sullivan2001Truth-tellingDiagnoses}
\begin{barticle}
\bauthor{\bsnm{Sullivan}, \binits{R.J.}},
\bauthor{\bsnm{Menapace}, \binits{L.W.}},
\bauthor{\bsnm{White}, \binits{R.M.}}:
\batitle{{Truth-telling and patient diagnoses}}.
\bjtitle{Journal of Medical Ethics}
\bvolume{27}(\bissue{3}),
\bfpage{192}--\blpage{197}
(\byear{2001})
\doiurl{10.1136/jme.27.3.192}
\end{barticle}
\endbibitem

\bibitem[\protect\citeauthoryear{Liley et~al.}{2023}]{Liley2023DevelopmentScotland}
\begin{barticle}
\bauthor{\bsnm{Liley}, \binits{J.}},
\bauthor{\bsnm{Bohner}, \binits{G.}},
\bauthor{\bsnm{Emerson}, \binits{S.R.}},
\bauthor{\bsnm{Mateen}, \binits{B.A.}},
\bauthor{\bsnm{Borland}, \binits{K.}},
\bauthor{\bsnm{Carr}, \binits{D.}},
\bauthor{\bsnm{Heald}, \binits{S.}},
\bauthor{\bsnm{Oduro}, \binits{S.D.}},
\bauthor{\bsnm{Ireland}, \binits{J.}},
\bauthor{\bsnm{Moffat}, \binits{K.}},
\bauthor{\bsnm{Porteous}, \binits{R.}},
\bauthor{\bsnm{Riddell}, \binits{S.}},
\bauthor{\bsnm{Cunningham}, \binits{N.}},
\bauthor{\bsnm{Holmes}, \binits{C.}},
\bauthor{\bsnm{Payne}, \binits{K.}},
\bauthor{\bsnm{Vollmer}, \binits{S.J.}},
\bauthor{\bsnm{Vallejos}, \binits{C.A.}},
\bauthor{\bsnm{Aslett}, \binits{L.J.M.}}:
\batitle{{Development and assessment of a machine learning tool for predicting emergency admission in Scotland}}.
\bjtitle{medRxiv}
(\byear{2023})
\doiurl{10.1101/2021.08.06.21261593}
\end{barticle}
\endbibitem

\bibitem[\protect\citeauthoryear{Wong et~al.}{2021}]{Wong2021ExternalPatients}
\begin{barticle}
\bauthor{\bsnm{Wong}, \binits{A.}},
\bauthor{\bsnm{Otles}, \binits{E.}},
\bauthor{\bsnm{Donnelly}, \binits{J.P.}},
\bauthor{\bsnm{Krumm}, \binits{A.}},
\bauthor{\bsnm{McCullough}, \binits{J.}},
\bauthor{\bsnm{DeTroyer-Cooley}, \binits{O.}},
\bauthor{\bsnm{Pestrue}, \binits{J.}},
\bauthor{\bsnm{Phillips}, \binits{M.}},
\bauthor{\bsnm{Konye}, \binits{J.}},
\bauthor{\bsnm{Penoza}, \binits{C.}},
\bauthor{\bsnm{Ghous}, \binits{M.}},
\bauthor{\bsnm{Singh}, \binits{K.}}:
\batitle{{External validation of a widely implemented proprietary sepsis prediction model in hospitalized patients}}.
\bjtitle{JAMA Internal Medicine}
\bvolume{181}(\bissue{8}),
\bfpage{1065}--\blpage{1070}
(\byear{2021})
\doiurl{10.1001/jamainternmed.2021.2626}
\end{barticle}
\endbibitem

\bibitem[\protect\citeauthoryear{Staffa and Zurakowski}{2021}]{Staffa2021StatisticalModels}
\begin{barticle}
\bauthor{\bsnm{Staffa}, \binits{S.J.}},
\bauthor{\bsnm{Zurakowski}, \binits{D.}}:
\batitle{{Statistical Development and Validation of Clinical Prediction Models}}.
\bjtitle{Anesthesiology}
\bvolume{135}(\bissue{3}),
\bfpage{396}--\blpage{405}
(\byear{2021})
\doiurl{10.1097/ALN.0000000000003871}
\end{barticle}
\endbibitem

\bibitem[\protect\citeauthoryear{Snooks et~al.}{2019}]{Snooks2019EffectsTrial}
\begin{barticle}
\bauthor{\bsnm{Snooks}, \binits{H.}},
\bauthor{\bsnm{Bailey-Jones}, \binits{K.}},
\bauthor{\bsnm{Burge-Jones}, \binits{D.}},
\bauthor{\bsnm{Dale}, \binits{J.}},
\bauthor{\bsnm{Davies}, \binits{J.}},
\bauthor{\bsnm{Evans}, \binits{B.A.}},
\bauthor{\bsnm{Farr}, \binits{A.}},
\bauthor{\bsnm{Fitzsimmons}, \binits{D.}},
\bauthor{\bsnm{Heaven}, \binits{M.}},
\bauthor{\bsnm{Howson}, \binits{H.}},
\bauthor{\bsnm{Hutchings}, \binits{H.}},
\bauthor{\bsnm{John}, \binits{G.}},
\bauthor{\bsnm{Kingston}, \binits{M.}},
\bauthor{\bsnm{Lewis}, \binits{L.}},
\bauthor{\bsnm{Phillips}, \binits{C.}},
\bauthor{\bsnm{Porter}, \binits{A.}},
\bauthor{\bsnm{Sewell}, \binits{B.}},
\bauthor{\bsnm{Warm}, \binits{D.}},
\bauthor{\bsnm{Watkins}, \binits{A.}},
\bauthor{\bsnm{Whitman}, \binits{S.}},
\bauthor{\bsnm{Williams}, \binits{V.}},
\bauthor{\bsnm{Russell}, \binits{I.}}:
\batitle{{Effects and costs of implementing predictive risk stratification in primary care: A randomised stepped wedge trial}}.
\bjtitle{BMJ Quality and Safety}
\bvolume{28}(\bissue{9}),
\bfpage{697}--\blpage{705}
(\byear{2019})
\doiurl{10.1136/bmjqs-2018-007976}
\end{barticle}
\endbibitem

\end{thebibliography}

\end{document}